\documentclass[conference,10pt]{IEEEtran}
\IEEEoverridecommandlockouts
\usepackage{amsmath}
\usepackage{iftex}
\usepackage[T1]{fontenc}
\ifPDFTeX
  \usepackage[utf8]{inputenc}
\fi
\usepackage{newtxtext,newtxmath}
\usepackage[varqu,scaled=0.92]{zi4}
\ifPDFTeX\else
\fi
\usepackage[american]{babel}
\usepackage{booktabs,array,tabularx}
\usepackage{graphicx}
\usepackage{newfloat}
\DeclareFloatingEnvironment[name=Listing,placement=!t]{listing}
\usepackage{microtype}
\usepackage{xcolor}
\usepackage{cite}
\usepackage[hidelinks]{hyperref}
\usepackage{xurl}
\hypersetup{pdftitle={Consistent LLM Call Handling Across Java Frameworks: A Reference Implementation and Benchmark},pdfauthor={Oleksandr Klymenko}}

\begin{document}

\title{Consistent LLM Call Handling Across Java Frameworks: A Reference Implementation and Benchmark}

\author{\IEEEauthorblockN{Oleksandr Klymenko}
\IEEEauthorblockA{Independent Researcher, Poland \\
ORCID 0009-0000-5924-3697 \\
alexanderklmn@gmail.com}}

\maketitle

\begin{abstract}
Java applications reach large language models (LLMs) through an integration framework, and rely on it to handle a failing provider: to retry what can be retried and nothing else, to wait as long as the provider asks, never to present a truncated answer as complete, and to record each failure in the application's metrics. We state these expectations as a six-clause reliability contract and measure Spring AI 2.0.1 and LangChain4j 1.21.0 against it with a scripted OpenAI-compatible provider: fifteen failure scenarios, synchronous and streaming, three repetitions. LangChain4j meets it in 47 of 77 applicable cases with its defaults and in 51 of 83 when configured explicitly, Spring AI in 72 of 77 and 76 of 83; both deliver a stream that ended without its terminator as a complete answer. We then place one implementation of the contract beneath both frameworks, at the HTTP boundary --- an OkHttp interceptor for Spring AI, an \texttt{HttpClient} for LangChain4j, and a \texttt{java.net.http} transport for code without a framework --- sharing a 725-line core that depends only on the JDK and the Micrometer Observation API. With it, all three clients behave identically and meet the contract in 90 of 91 cases; the remaining case is a policy trade-off between retrying a hung provider and waiting for a slow first token. Against a local model server the layer rejected none of 700 calls, and it adds a median of 32--217 $\mu$s to a call.
\end{abstract}


\section{Introduction}

A Java application that calls a large language model (LLM) rarely speaks HTTP to the provider itself. It uses an integration framework --- Spring AI or LangChain4j --- which turns a prompt into a request, the response into an object, and a failure into an exception, a retry, or nothing. Which of the three happens decides what the application and its operator see. A rate limit absorbed by a retry is invisible; a stream that ended early and is delivered as a complete answer is indistinguishable from a real one; a failure that no metric records does not appear on a dashboard. Such failures are \emph{gray} in the sense of~\cite{r1}: the component fails, and the observer that should notice does not.

Providers fail in ordinary, documented ways. They answer 429 with a \texttt{Retry-After}, return 5xx under load, stop responding, close connections, and end streams early. HTTP semantics~\cite{r2,r3} and the provider's streaming format~\cite{r4} say what a client should do in most of these cases. The frameworks implement these rules separately, each in its own retry loop, timeout model and instrumentation, and they implement them differently. Section V.B measures the difference: run against the same fifteen scripted provider failures, LangChain4j 1.21.0 meets a six-clause reliability contract in 47 of 77 applicable cases with its defaults and Spring AI 2.0.1 in 72 of 77, and the two fail on different clauses.

This paper argues that the rules belong below the frameworks rather than inside each of them. Both frameworks reach an OpenAI-compatible endpoint over HTTP, and both let the application supply the HTTP client: Spring AI through an OkHttp interceptor, LangChain4j through its \texttt{HttpClient} interface. Beneath that seam the wire is the same --- a JSON body or a stream of server-sent events --- so a layer placed there can apply one contract to every framework above it, and to an application that uses no framework at all.

We contribute:

\begin{enumerate}
\item \textbf{A reliability contract} for model calls in six clauses (Section III), each derived from HTTP semantics, the streaming format, or the observability requirement, and each checkable against a scripted provider.
\item \textbf{A layer that implements it once} (Section IV): a framework-independent core of 725 lines with no dependency beyond the JDK and the Micrometer Observation API, and three adapters --- an OkHttp interceptor for Spring AI (210 lines), a LangChain4j \texttt{HttpClient} (137 lines), and a \texttt{java.net.http} transport (290 lines) usable without a framework.
\item \textbf{An evaluation} (Section V) by fault injection --- fifteen scenarios, synchronous and streaming, three repetitions, before and after --- against a real model server (700 calls to two local models), and of the latency the layer adds.
\item \textbf{An open implementation} with every result reproducible from one command~\cite{r5}.
\end{enumerate}

\section{Background and related work}

\textbf{Two frameworks, two retry mechanisms.} Spring AI 2.0~\cite{r6} reaches OpenAI through the official \texttt{openai-java} SDK~\cite{r7}; retries, back-off and \texttt{Retry-After} handling are the SDK's, configured by Spring AI's \texttt{maxRetries} (default 3) and \texttt{timeout} (default 60~s, a bound on the whole call), and Spring AI instruments the call with Micrometer observations~\cite{r8}. LangChain4j 1.21.0~\cite{r9} has its own HTTP client abstraction and its own retry loop for synchronous calls (default \texttt{maxRetries} 2); its instrumentation is a listener interface, and the module \texttt{langchain4j-micrometer-metrics} records Micrometer meters from it. The streaming path of LangChain4j's OpenAI client has no retry setting.

\textbf{What a client should do.} A 5xx response is a server error and may succeed on repetition~\cite[\S15.6]{r2}; a 4xx response is a client error, and for 400 or 401 the same request will fail again~\cite[\S15.5]{r2}; a 429 asks the client to slow down, and \texttt{Retry-After} says for how long~\cite[\S10.2.3]{r2},~\cite[\S4]{r3}. A model request is a POST: a request that reached the provider may have been executed and billed even if the client never saw the answer, and repeating a request whose \emph{successful} response could not be decoded repeats a completed generation. An OpenAI chat-completions stream ends with a chunk that carries a \texttt{finish\_reason} and a final \texttt{data: [DONE]}~\cite{r4}; a \texttt{data:} event whose JSON carries an \texttt{error} object reports a failure~\cite{r7}. A stream that ends without its terminator is incomplete.

\textbf{Retries and timeouts as a source of failure.} Studies of cloud-system failures find timeout and retry handling among the recurring causes of incidents~\cite{r10,r11,r12,r13}: missing or misconfigured timeouts, retries that amplify load, and errors that are swallowed rather than reported. Exponential back-off goes back to Ethernet~\cite{r14}; randomising it spreads retries in time; and at scale, repeated requests are both a remedy for tail latency and a source of extra load~\cite{r15}. Grottke and Trivedi classify retry as one of four responses to software faults~\cite{r16}. Fault injection into a running system to observe its handling of failure is the method of chaos engineering~\cite{r17}; we use it here at the scale of one client and a scripted provider.

\textbf{Resilience libraries and gateways.} General-purpose resilience libraries for the JVM --- Resilience4j, Spring Retry, Failsafe --- provide retry, back-off, time-limit and circuit-breaker policies around any call~\cite{r18,r19,r20}. They decide on the exception a call throws, after the HTTP response has been turned into an object, and so cannot see what C3 and C4 require: whether a stream ended with its terminator, or how long the provider has been silent between events. Wrapping a framework call in such a policy also nests a second retry loop around the framework's own. LLM gateways --- proxies such as LiteLLM~\cite{r21} --- apply retries and fallbacks between the application and the provider, but as a separate network service, and the application's own metrics then see only the gateway. The layer proposed here is a library at the HTTP boundary of the process: it sees the wire, as a gateway does, and records in the application's own observations, as a library does.

\textbf{Developers of LLM applications.} An analysis of developer questions about LLM APIs finds error handling, rate limits and response parsing among the recurring difficulties~\cite{r22}. The present paper takes one class of these --- provider failures on the HTTP path --- and moves their handling out of application code and out of each framework into one layer.

\section{The contract}

We call one invocation of the framework's API a \emph{call}, and each HTTP request it causes an \emph{attempt}. An attempt fails with a \emph{kind} $f$ from a fixed set $F$: rate-limited (429), server error (5xx), request timeout (408), connection, no response, authentication (401, 403), client error (other 4xx), malformed response, empty response, truncated stream, stream error event, idle stream, cancelled. A subset $F_T\subset F$ is \emph{transient} --- the first five --- and the rest are permanent. The contract has six clauses.

\textbf{C1 --- retry exactly the transient failures.} After attempt $k$ fails with kind $f$, another attempt is made if and only if

\begin{equation}
\begin{aligned} R(k,f)\;=\;&\big[f\in F_T\big]\;\wedge\;\neg\delta\;\wedge\;k<K\\ &\wedge\;\big[a\le A_{\max}\big]\;\wedge\;\big[e_k+w_k+\tau\le B\big], \end{aligned}
\label{eq:1}
\end{equation}

\noindent where $\delta$ states that part of the response was already delivered to the caller, $K$ is the attempt limit, $a$ the delay the provider asked for (if any), $A_{\max}$ the longest such delay the client will honour, $e_k$ the time elapsed since the call began, $w_k$ the wait before attempt $k+1$, $\tau$ the next attempt's timeout and $B$ the call's total budget. The clause forbids two errors: not retrying a transient failure, and retrying a permanent one --- in particular, repeating a request whose successful response was malformed.

\textbf{C2 --- wait as long as the provider asks.} If the provider states a delay $a$ (\texttt{retry-after-ms}, or \texttt{Retry-After} in seconds or as a date), the wait is $w_k=a$; a delay above $A_{\max}$ ends the call, because waiting less than asked only provokes another refusal. Otherwise the wait doubles from a base $b$ up to a cap $c$, with a fraction $j$ drawn at random~\cite{r14,r15}:

\begin{equation}
w_k \;=\; \min\!\big(c,\; b\,2^{\,k-1}\big)\,\big(1-j+j\,U_k\big),\qquad U_k\sim\mathcal U[0,1].
\label{eq:2}
\end{equation}

\textbf{C3 --- never present an incomplete answer as complete.} The application receives either the complete answer or an exception. A non-streaming body must be one JSON object, and if it carries \texttt{choices}, at least one. A stream is complete if and only if it contains a terminal event --- \texttt{data: [DONE]} or a chunk with a non-null \texttt{finish\_reason}; an error event or an event that is not JSON makes it fail. Events received before a failure are delivered before the exception, so that the application can keep or discard a partial answer knowingly.

\textbf{C4 --- bound silence, not length.} A non-streaming attempt is bounded by a response timeout $\tau_r$; a streaming attempt by a first-event timeout $\tau_1$ and, once events flow, by an idle timeout $\tau_i$ between events. A single bound on the whole call conflates a provider that is silent with one that is slowly producing a long answer. The total budget $B$ bounds the time until the response starts:

\begin{equation}
T_{\text{start}} \;\le\; B,
\label{eq:3}
\end{equation}

\noindent since by (1) no attempt starts unless it can end within $B$, and each attempt's timeout is $\min(\tau,\,B-e_k)$. A stream that has started is bounded by $\tau_i$ per gap, not by $B$.

\textbf{C5 --- record every outcome as it was.} Each call ends in the instrumentation in exactly one of two states, success or failure, matching what the application received, and a failure carries its kind --- not the class of a wrapper exception. This includes a truncated stream, a stream that failed after events were delivered, and a stream the caller abandoned.

\textbf{C6 --- make every attempt visible.} The instrumentation holds one record per request that reached the provider, with its status, kind and the retry decision. Retries that only the server log can count are invisible to the operator.

The clauses are not new individually; each framework implements some of them. What is new is stating them together, as one contract that a layer either meets or does not, and checking them per scenario (Section V).

\section{Design}

\subsection{Placement}

\begin{figure*}[!t]
\centering
\includegraphics[width=0.86\textwidth]{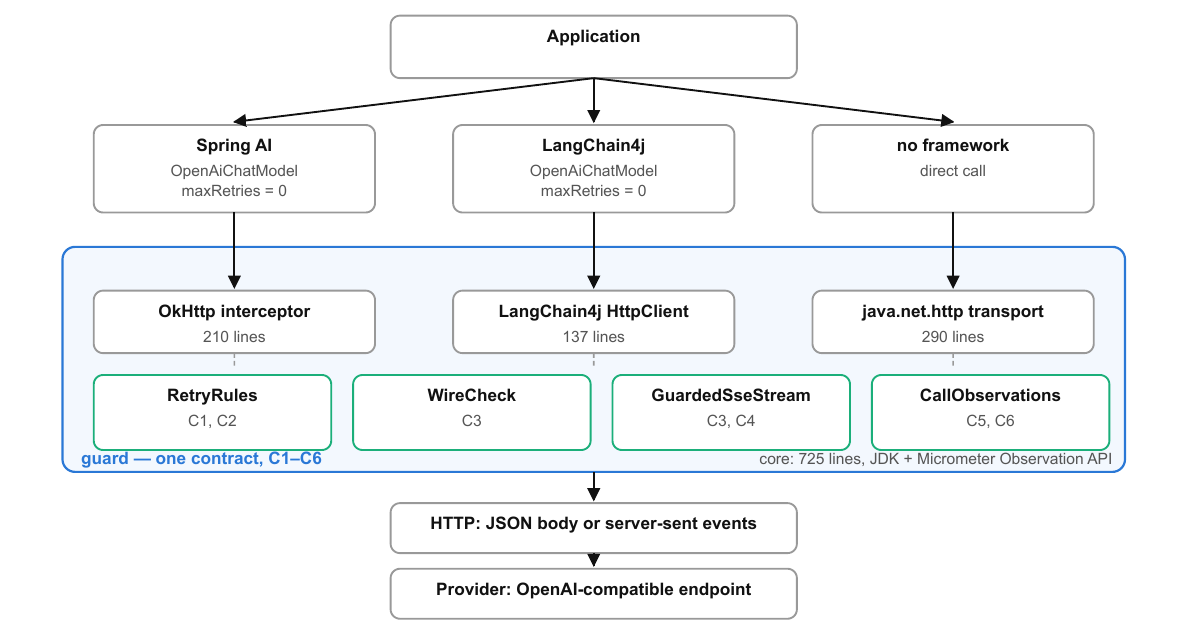}
\caption{Placement of the layer. Each framework keeps its API and its exception types; its own retries are switched off and its HTTP client is replaced by an adapter that delegates to one core implementing the contract. Code without a framework uses the \texttt{java.net.http} transport directly.}
\label{fig:1}
\end{figure*}

The layer sits at the HTTP boundary, beneath the framework and above the socket (Fig.~\ref{fig:1}). Three properties of that place make one implementation sufficient. First, the frameworks expose it: Spring AI accepts OkHttp application interceptors through \texttt{httpClientBuilderCustomizer}, and LangChain4j accepts any implementation of its \texttt{HttpClient} interface through \texttt{httpClientBuilder}. Second, everything the contract needs is visible there --- status, headers, body bytes and their timing --- and nothing it needs is visible only higher up. Third, the wire is common: every OpenAI-compatible endpoint --- OpenAI, Azure OpenAI, Ollama, vLLM --- speaks the same JSON and server-sent-event format.

The price of the placement is that the framework must step aside: its own retries are set to zero, and its own call timeout above the layer's budget, so that one component decides. Final error responses (a 400, or a 503 after the last attempt) are passed up unchanged, so that each framework still maps them to its own exception types; failures that only the layer detects --- no response, a malformed or empty answer, a truncated stream, an error event, silence --- are raised as a \texttt{GuardException} that carries the kind.

\subsection{Components}

\textbf{Retry rules.} One function implements (1)--(2) and is called by every transport, so the decision is identical whichever framework is above. The kind of a failure is the only input that differs between transports: an HTTP status maps to a kind by its class; a read timeout before the response maps to \emph{no response}; an I/O failure before any byte maps to \emph{connection}.

\textbf{Wire check.} A non-streaming 2xx body is parsed before it is handed up; a malformed body or an empty \texttt{choices} array ends the call with a permanent kind. The parser is a 170-line reader written for the purpose, so that the core carries no JSON library and cannot conflict with the framework's.

\textbf{Guarded event stream.} A streaming response is wrapped in an input stream that passes bytes to the caller one \emph{event} at a time, and only after the event has been classified. The design rests on two properties, both enforced by construction and tested on every split of a stream into chunks:

\emph{Proposition 1 (completeness).} The guarded stream reports its end to the caller if and only if a terminal event has been received; otherwise the read that reaches the end of the body throws \emph{truncated}. \emph{Proposition 2 (ordering).} The bytes of an error event or of a malformed event are never delivered; every event received before it is delivered before the exception.

Before the first event the stream is not yet handed to the caller, so a silence or a broken connection at that point is still a transient failure of the attempt and may be retried; after it, $\delta$ in (1) is true and nothing is retried.

\textbf{Observations.} Each call is one Micrometer observation \texttt{llm.guard.call}, with low-cardinality keys \texttt{transport}, \texttt{mode}, \texttt{outcome}, \texttt{error.type} (the kind) and \texttt{attempts}; each attempt is a child observation \texttt{llm.guard.attempt} with \texttt{status}, \texttt{error.type} and \texttt{decision} (retry or stop). Any observation handler --- a meter registry, a tracer --- receives them. A stream's call observation ends when the stream ends, fails, or is closed early by the caller.

\subsection{Adapters}

\textbf{Spring AI (OkHttp interceptor, 210 lines).} The interceptor runs each attempt through \texttt{chain.proceed}, with the attempt's timeout set as the chain's read timeout; for a stream it reads until the first event before returning the response, and wraps the rest of the body in the guarded stream. Retries happen inside the interceptor, beneath the SDK's own (disabled) retry loop.

\textbf{LangChain4j (\texttt{HttpClient}, 137 lines).} The adapter implements LangChain4j's synchronous \texttt{execute} and its server-sent-event \texttt{execute}, maps a final error status to LangChain4j's \texttt{HttpException} and \emph{no response} to its \texttt{TimeoutException}, so that the model's own exception mapping is unchanged, and follows the same callback sequence as LangChain4j's JDK client (open, parse, close).

\textbf{No framework (\texttt{java.net.http}, 290 lines).} The JDK client has no read timeout for a body once its headers have arrived, so the transport reads the body on a virtual thread into a queue that the consumer polls with the idle timeout.

Installing the layer is a builder argument in each framework (Listing~\ref{lst:1}).

\begin{listing*}[!t]
\footnotesize
\begin{verbatim}
// Spring AI
var guard = new GuardInterceptor(Policy.defaults(), registry);
var model = OpenAiChatModel.builder()
    .options(OpenAiChatOptions.builder().model(m).maxRetries(0).timeout(Duration.ofMinutes(10)).build())
    .httpClientBuilderCustomizer(b -> b.interceptor(guard)).build();
// LangChain4j
var model = OpenAiChatModel.builder().modelName(m).maxRetries(0)
    .httpClientBuilder(new GuardedHttpClientBuilder(Policy.defaults(), registry)).build();
\end{verbatim}
\vspace{-4pt}
\caption{Installing the layer beneath Spring AI and beneath LangChain4j.}
\label{lst:1}
\end{listing*}

\section{Evaluation}

\subsection{Setup}

\textbf{Scripted provider.} A local OpenAI-compatible endpoint whose behaviour is scripted per attempt and which records the arrival time of every request, so that attempts and the gaps between them are measured at the server, independently of what the client reports. Fifteen scenarios cover server errors (S02, S03), rate limiting with \texttt{Retry-After: 3} (S04, S05), client errors (S06, S07), a provider that does not answer (S08: 5~s once; S14: 70~s on every attempt), a 2xx body that is not JSON (S09) or has an empty \texttt{choices} array (S10), a stream closed after three of eight chunks without its terminator (S11), a stream carrying an error event after three chunks (S12), and slow but live streams (S13: 1.5~s between chunks; S15: 8.5~s between chunks, 68~s in all).

\textbf{Clients and setups.} Spring AI 2.0.1 (\texttt{openai-java} 4.49.0) and LangChain4j 1.21.0, each in synchronous and streaming mode, each in three setups: its \emph{defaults}; an \emph{explicit} configuration (\texttt{maxRetries} 2, timeout 2~s); and with the \emph{guard} installed (framework retries 0, framework timeout 10 min). The guard is also run alone over \texttt{java.net.http} (\emph{plain}). The guard's policy in the study is $K=3$, $\tau_r=\tau_1=3$ s, $\tau_i=10$ s, $B=60$ s, $b=0.5$ s, $c=8$ s, $A_{\max}=30$ s, $j=0.5$, with a fixed random seed; the short timeouts suit a provider that answers at once. The \emph{explicit} configuration is the one a developer would set for such a provider --- two retries and a 2~s timeout, close to the guard's 3~s --- so that the frameworks are compared with the guard at a similar strictness, and not only at their more lenient defaults.

\textbf{Scoring.} A \emph{cell} is one scenario $\times$ client $\times$ mode $\times$ setup; it meets a clause only if all three repetitions do (S14 and S15, which wait for timeouts, run once). For C5 a metric counts as recording a failure if, for Spring AI, its \texttt{gen\_ai.client.operation} observation stops with an error; for LangChain4j, a meter carries an error tag (its listener callback alone is not a metric); for the guard, \texttt{llm.guard.call} ends as failure --- and, where the scenario determines it, with the expected kind. Clauses that do not apply to a scenario are not counted, so the denominators differ between columns. With $\mathcal C$ the cells of a client and setup, $A_c(k)$ the clauses applicable in cell $k$ and $m_{c,k,r}\in\{0,1\}$ whether repetition $r$ of cell $k$ meets clause $c$, the score reported is

\begin{equation}
\frac{\sum_{k\in\mathcal C}\sum_{c\in A_c(k)} \prod_{r} m_{c,k,r}}{\sum_{k\in\mathcal C}\,|A_c(k)|}.
\label{eq:4}
\end{equation}

\textbf{Real server and overhead.} The guard was run against Ollama 0.34.2~\cite{r23} on one RTX 3090 with \texttt{llama3.2} (3~B parameters; 50 calls per cell) and \texttt{qwen2.5:7b} (20 per cell), all clients, both modes, without and with the guard, with \texttt{Policy.defaults()} ($\tau_r=60$ s, $\tau_1=\tau_i=30$ s, $B=5$ min). Overhead was measured as the latency of 2,000 successful calls per configuration against the scripted provider, with the same client built once with and once without the guard, in alternating blocks of 50 after 300 warm-up calls.

\subsection{Contract compliance, before and after}

Table~\ref{tab:1} gives, for each client and setup, the number of cells that meet each clause.

\begin{table*}[!t]
\centering
\footnotesize
\setlength{\tabcolsep}{4pt}
\caption{Cells meeting each clause / cells to which it applies (all repetitions agree).}
\label{tab:1}
\begin{tabularx}{\textwidth}{>{\raggedright\arraybackslash}p{3.1cm}>{\centering\arraybackslash}X>{\centering\arraybackslash}X>{\centering\arraybackslash}X>{\centering\arraybackslash}X>{\centering\arraybackslash}X>{\centering\arraybackslash}X>{\centering\arraybackslash}X}
\toprule
\textbf{Clause} & \textbf{LC4j defaults} & \textbf{LC4j explicit} & \textbf{LC4j + guard} & \textbf{Spring AI defaults} & \textbf{Spring AI explicit} & \textbf{Spring AI + guard} & \textbf{plain + guard} \\
\midrule
C1 retry exactly the transient & 11/15 & 11/16 & 17/17 & 15/15 & 15/16 & 17/17 & 17/17 \\
C2 wait as asked & 0/3 & 0/3 & 3/3 & 3/3 & 3/3 & 3/3 & 3/3 \\
C3 no incomplete answer as complete & 16/18 & 17/19 & 21/21 & 16/18 & 17/19 & 21/21 & 21/21 \\
C4 bound silence, not length & 3/5 & 4/7 & 7/8 & 4/5 & 5/7 & 7/8 & 7/8 \\
C5 record every outcome & 5/18 & 6/19 & 21/21 & 16/18 & 17/19 & 21/21 & 21/21 \\
C6 every attempt visible & 12/18 & 13/19 & 21/21 & 18/18 & 19/19 & 21/21 & 21/21 \\
\textbf{all} & \textbf{47/77} & \textbf{51/83} & \textbf{90/91} & \textbf{72/77} & \textbf{76/83} & \textbf{90/91} & \textbf{90/91} \\
\bottomrule
\end{tabularx}
\end{table*}

\begin{figure*}[!t]
\centering
\includegraphics[width=0.86\textwidth]{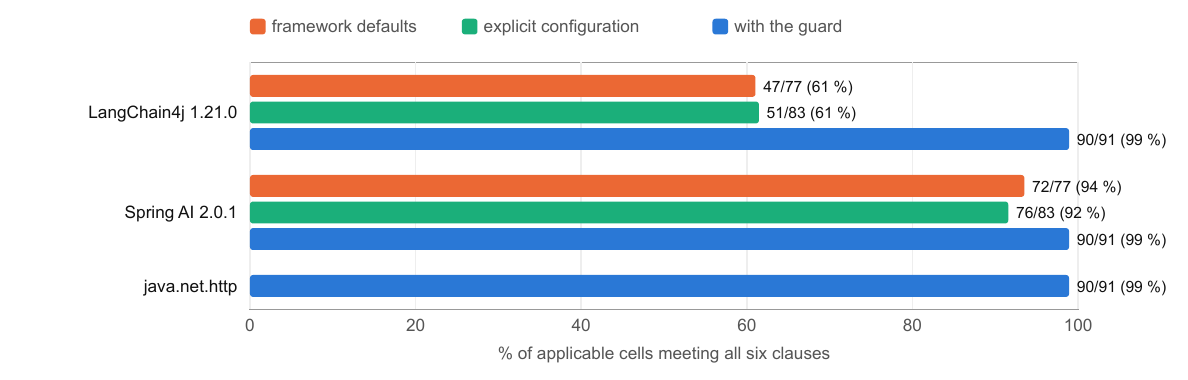}
\caption{Share of applicable cells that meet all six clauses (Table~\ref{tab:1}, last row): each framework with its defaults, with an explicit configuration, and with the guard installed; and the guard without a framework.}
\label{fig:2}
\end{figure*}

Without the guard, the two frameworks differ more from each other than from the contract. LangChain4j's streaming path makes no retry, so a single 500 (S02), a 429 (S04) or a 5~s silence (S08) reaches the application as an error (C1, C4); its synchronous path retries a 429 after its own back-off of about 0.5~s instead of the 3~s asked for (minimum gap 507--539~ms; C2), and repeats a request whose 200 response was not JSON (S09: three attempts; C1). Its Micrometer listener records no meter for any failure (C5: 5 of 18 cells), and its listener sees one request where the provider received three (C6). Spring AI meets C1, C2 and C6 in every applicable cell except one --- an asynchronous streaming call that timed out is not retried (S08) --- and records failures as observation errors. Both frameworks deliver a stream that ended without its terminator (S11) as a complete answer and record it as a success (C3, C5); LangChain4j does the same with a stream that carried an error event (S12); Spring AI returns an empty \texttt{choices} array as an empty answer recorded as a success (S10). Spring AI's timeout bounds the whole call, so a stream that is still delivering is cut off at 2~s (S13, explicit) or at 60~s (S15, defaults: seven of eight chunks delivered, then an error).

With the guard, the three clients behave identically in every scenario: the same attempts, the same gaps (36 Retry-After waits of 3,001--3,007~ms, median 3,002~ms, for \texttt{Retry-After: 3}), the same kind recorded, and an exception --- with the delivered chunks kept --- for S10, S11 and S12. The exception reaching the application still differs by framework (Spring AI wraps the guard's failure in \texttt{OpenAIIoException}, LangChain4j in its \texttt{TimeoutException} or a runtime exception), but the kind is in the cause chain and in the observation.

\textbf{The one cell the guard misses} in every client is S15, a stream whose first chunk arrives after 8.5~s: with $\tau_1=3$ s the guard treats the silence as \emph{no response}, retries twice and fails after 10.2--10.3~s. This is the policy, not a defect: the study's $\tau_1$ is chosen below 5~s so that the provider that hangs for 5~s in S08 is retried, and no single value of $\tau_1$ can both retry a 5~s silence and wait out an 8.5~s one --- before the first byte, a hung provider and a slow first token look the same. Run with $\tau_1=10$ s, S15 completes on all three clients in one attempt (\texttt{results/outcomes\_S15\_firstEvent10s.csv}), which shows that once a stream has started, only the 10~s idle bound applies, not its 68~s length.

\textbf{Bounded failure time.} A provider that never answers (S14) costs a synchronous call 243~s with Spring AI's defaults (four attempts of 60~s) and 181~s with LangChain4j's (three of 60~s); with the guard, 10.15--10.16~s for three attempts on every client, inside the budget as (3) requires.

\subsection{Real server}

Against Ollama, all 700 calls succeeded in every client, mode and setup (Table~\ref{tab:2}): the guard rejected no response and no stream from a server that was working, and its call observation recorded every guarded call as a success in one attempt. Median latencies with and without the guard differ by less than the variation of generation time between calls.

\begin{table*}[!t]
\centering
\footnotesize
\setlength{\tabcolsep}{4pt}
\caption{Real server: successful calls / calls, and median latency (ms).}
\label{tab:2}
\begin{tabularx}{\textwidth}{>{\raggedright\arraybackslash}p{3.1cm}>{\centering\arraybackslash}X>{\centering\arraybackslash}X>{\centering\arraybackslash}X>{\centering\arraybackslash}X>{\centering\arraybackslash}X}
\toprule
\textbf{Client} & \textbf{Mode} & \textbf{llama3.2 native} & \textbf{llama3.2 + guard} & \textbf{qwen2.5:7b native} & \textbf{qwen2.5:7b + guard} \\
\midrule
Spring AI & sync & 50/50 \textperiodcentered{} 229 & 50/50 \textperiodcentered{} 221 & 20/20 \textperiodcentered{} 342 & 20/20 \textperiodcentered{} 391 \\
Spring AI & stream & 50/50 \textperiodcentered{} 223 & 50/50 \textperiodcentered{} 234 & 20/20 \textperiodcentered{} 380 & 20/20 \textperiodcentered{} 327 \\
LangChain4j & sync & 50/50 \textperiodcentered{} 241 & 50/50 \textperiodcentered{} 236 & 20/20 \textperiodcentered{} 353 & 20/20 \textperiodcentered{} 342 \\
LangChain4j & stream & 50/50 \textperiodcentered{} 224 & 50/50 \textperiodcentered{} 245 & 20/20 \textperiodcentered{} 363 & 20/20 \textperiodcentered{} 351 \\
plain & sync & --- & 50/50 \textperiodcentered{} 243 & --- & 20/20 \textperiodcentered{} 361 \\
plain & stream & --- & 50/50 \textperiodcentered{} 239 & --- & 20/20 \textperiodcentered{} 370 \\
\bottomrule
\end{tabularx}
\end{table*}

\subsection{Overhead and size}

\begin{table}[!t]
\centering
\footnotesize
\setlength{\tabcolsep}{4pt}
\caption{Latency added to a successful call by the guard, against the scripted provider (2,000 calls each).}
\label{tab:3}
\begin{tabularx}{\columnwidth}{l>{\centering\arraybackslash}X>{\centering\arraybackslash}X>{\centering\arraybackslash}X>{\centering\arraybackslash}X}
\toprule
\textbf{Client} & \textbf{Mode} & \textbf{Median without / with ($\mu$s)} & \textbf{Added, median ($\mu$s)} & \textbf{Added, P95 ($\mu$s)} \\
\midrule
Spring AI & sync & 582 / 645 & 63 & 987 \\
Spring AI & stream & 839 / 977 & 138 & 576 \\
LangChain4j & sync & 621 / 653 & 32 & 201 \\
LangChain4j & stream & 522 / 739 & 217 & 617 \\
java.net.http & sync & 325 / 364 & 39 & 90 \\
java.net.http & stream & 358 / 513 & 155 & 313 \\
\bottomrule
\end{tabularx}
\end{table}

The median cost is 32--217 $\mu$s per call (Table~\ref{tab:3}), larger for streams, where every event is parsed before it is passed on. Against generation times of 220--390~ms for the smallest local models (Table~\ref{tab:2}) this is below 0.1~\%. The core is 725 non-comment lines; the adapters are 210 (OkHttp), 137 (LangChain4j) and 290 (\texttt{java.net.http}, including the body pump). The tests cover the retry rules, \texttt{Retry-After} parsing, the wire checks and the stream gate on every split of a stream into chunks (16 tests).

\section{Discussion}

\textbf{Where it is used.} The layer serves in this role in~\cite{r24}, an operator-facing explainer for the fault attributions of~\cite{r25}: there, local models are reached through both Spring AI and LangChain4j with the guard installed, and a failing model server degrades the operator's explanation to a template within the guard's budget instead of stalling the triage.

\textbf{What the placement buys.} One implementation of the contract served three clients with identical behaviour, and the difference between Spring AI with the guard and LangChain4j with the guard vanished in every measured respect except the class of the exception. Fixing each deviation in each framework is also needed, and several of the deviations measured here are suitable for reports to the projects; but an application cannot wait for every framework to converge, and the layer gives it the contract now and keeps it if the application changes framework.

\textbf{What it costs.} The frameworks must step aside: their retries off, their call timeout out of the way. Their own instrumentation still runs --- Spring AI's observation, LangChain4j's listener --- but sees one call where the guard made several. The OkHttp adapter waits for a retry on the thread that runs the interceptor, which on Spring AI's streaming path is an OkHttp dispatcher thread; under heavy concurrency the dispatcher's limits apply.

\textbf{Limits of the evaluation.} The scripted provider reproduces failures at the protocol level; it does not reproduce a provider's load behaviour, and the real-server runs exercise only the success path. One policy was evaluated, and S15 shows that the first-event timeout is a trade-off the operator must make knowingly. The contract covers the HTTP path of a single call --- not tool-calling loops, conversation state, or the content of an answer. And it covers the OpenAI-compatible wire only; other provider protocols need their own wire checks, although the retry rules, the stream gate and the observations carry over unchanged. The overhead was measured on one machine while a model server served another experiment on its GPU. The measurements were made on 6 October 2026 with Spring AI 2.0.1 and LangChain4j 1.21.0; later releases (LangChain4j 1.22.0 appeared on 8 October) are not covered, although the harness re-runs against any version.

\textbf{Whether to retry a timeout} remains a policy decision the layer cannot make for the application: a request that timed out may have been executed and billed. The guard retries it by default, as both frameworks do on their synchronous paths; $K=1$ turns this off.

\section{Conclusion}

The handling of provider failures in Java LLM applications is currently a property of the framework, and the two frameworks studied here handle the same failures differently: measured against a six-clause contract derived from HTTP semantics, the streaming format and the requirement that failures be visible, they meet it in 47--51 of 77--83 applicable cases (LangChain4j) and 72--76 (Spring AI). A layer placed beneath them, at the HTTP boundary, implements the contract once --- 725 lines of core and 137--290 lines per adapter --- and gives Spring AI, LangChain4j and plain \texttt{java.net.http} the same behaviour: 90 of 91 cells, the remaining one a policy trade-off between a hung provider and a slow first token. It rejected none of 700 calls to a working model server and adds 32--217 $\mu$s to a call. The contract, the layer and the evaluation are open, and the evaluation can be re-run against any later version of either framework.

\section*{Data and code availability}

The core, the three adapters, the scripted provider, the evaluation harness, every result file and the figure scripts are available at \url{https://github.com/alxkm/llm-call-guard} under the MIT license; one command re-runs the evaluation, and the repository records the runtime environment of each run.

\section*{Competing interests}

The author contributes to the open-source projects Spring AI and LangChain4j, whose behaviour is measured here --- among others, tests for the observation components of Spring AI (spring-projects/spring-ai\#4124, spring-projects/spring-ai\#4157, spring-projects/spring-ai\#4259, spring-projects/spring-ai\#4313) and a fix to token accounting in LangChain4j (langchain4j/langchain4j\#6322) --- and to Micrometer (micrometer-metrics/micrometer\#7924). These are tests and small fixes; none changes the behaviour measured here. The deviations reported are measured against a contract stated in advance, by a harness that is released in full. The work received no external funding.

\end{document}